\documentclass[]{spie}  

\usepackage{amsmath,amsfonts,amssymb}
\usepackage{graphicx}
\usepackage[colorlinks=true, allcolors=blue]{hyperref}
\usepackage{multirow}

\title{Mitigating annotation shift in cancer classification using single image generative models 
}

\vspace{0.25cm}
\author[a]{Marta Buetas Arcas}
\author[a, b, c]{Richard Osuala}
\author[a, d]{Karim Lekadir}
\author[a, e]{Oliver Díaz}
\vspace{-0.25cm}
\affil[a]{Departament de Matemàtiques i Informàtica, Universitat de Barcelona, Barcelona, Spain}

\affil[b]{Helmholtz Center Munich, Munich, Germany}

\affil[c]{Technical University of Munich, Munich, Germany}

\affil[d]{Institució Catalana de Recerca i Estudis Avançats (ICREA), Barcelona, Spain}

\affil[e]{Computer Vision Center, Bellaterra, Spain}

\begin{document} 
\maketitle

\section{ABSTRACT}
Artificial Intelligence (AI) has emerged as a valuable tool for assisting radiologists in breast cancer detection and diagnosis. However, the success of AI applications in this domain is restricted by the quantity and quality of available data, posing challenges due to limited and costly data annotation procedures that often lead to annotation shifts. 
This study simulates, analyses and mitigates annotation shifts in cancer classification in the breast mammography domain. First, a high-accuracy cancer risk prediction model is developed, which effectively distinguishes benign from malignant lesions. Next, model performance is used to quantify the impact of annotation shift. We uncover a substantial impact of annotation shift on multiclass classification performance particularly for malignant lesions. We thus propose a training data augmentation approach based on single-image generative models for the affected class, requiring as few as four in-domain annotations to considerably mitigate annotation shift, while also addressing dataset imbalance. 
Lastly, we further increase performance by proposing and validating an ensemble architecture based on multiple models trained under different data augmentation regimes. Our study offers key insights into annotation shift in deep learning breast cancer classification and explores the potential of single-image generative models to overcome domain shift challenges.
All code used for this study is made publicly available at \href{https://github.com/MartaBuetas/EnhancingBreastCancerDiagnosis/}{https://github.com/MartaBuetas/EnhancingBreastCancerDiagnosis}.

\keywords{Dataset Shift, Image Synthesis, Mammography, Synthetic Data, GANs, Deep Learning}

\section{INTRODUCTION} 
\label{sec:intro} 

Artificial Intelligence (AI) has emerged as a beneficial tool to assist radiologists 
in breast cancer detection and diagnosis~\cite{sharma2023multi,dembrower2023artificial, data_synthesis_review, salim2020external}. Both the quantity and quality of available data have a direct impact on the success of these applications. One major challenge is the scarcity of labeled data~\cite{data_synthesis_review, garrucho2023high}, largely attributable to the extensive time, effort, and costs associated with acquiring expert annotations. This often limits the training and evaluation of AI models, causing a lack of generalisation and robustness. This issue is further exacerbated by the varying quality of available expert annotations that commonly display high intra- and inter-observer variability~\cite{radiologist_variability}. This can lead to annotation shift~\cite{causality_matters}, where a model's performance decreases at test time if test annotations differ from their training counterparts, for example, in size, accuracy, delineation, lesion boundary and margin definition, annotation protocol or sourcing modality.
 
To increase classification model robustness against annotation shift, one approach is to generate additional training images that correspond to the desired annotation characteristics of the target domain~\cite{data_synthesis_review} (in-domain). To this end, we propose the selection of a single well-annotated in-domain training image to train a generative AI model, which, in turn, learns to synthesize an arbitrary number of variations. These variations are readily usable as additional classification model training images.

Following this approach, the contributions presented in this study are threefold: 
\vspace{-0.25cm} 
\begin{itemize}
\item Design and implementation of a high-accuracy malignancy classification model trained to distinguish cancerous from benign breast lesions. 
\vspace{-0.25cm} 
\item Identification and quantification 
of the impact of annotation shift on multiclass classification performance.

\vspace{-0.25cm} 
\item 
A novel investigation into
single-image generative AI models to mitigate annotation shift by requiring as few as only one additional annotation.
\vspace{-0.25cm} 
\end{itemize}

\section{METHODS AND MATERIALS}

The dataset used in this study is the Breast Cancer Digital Repository (BCDR)~\cite{bcdr_dataset}, an extensive accessible repository that comprises annotated cases of breast cancer patients from the northern region of Portugal. BCDR provides both normal and annotated patient cases, including digitised screen-film mammography and full-field digital mammography images. Each mammogram is accompanied by corresponding clinical data. The dataset consists on 984 mammograms with breast lesions. Among these, 386 are digitised screen-film and 598 digital images from 425 women. Each image is annotated with a binary mask delineated by radiologists to indicate regions-of-interest. Within this dataset, 699 cases are biopsy-proven benign, and 285 are malignant. Additionally, 200 normal mammograms from 48 women are included. To ensure the classifier's robustness for both formats, both film and digital images are equally included in all experiments.

The classification task in this study is to obtain a pre-biopsy cancer risk prediction of breast lesions, which can assist healthcare professionals in making informed decisions regarding further diagnostic procedures, interventions, and treatment plans. Our multiclass deep-learning classification model distinguishes between (i) healthy tissue, (ii) benign, and (iii) malignant lesions, thereby extending on previous approaches, which focused on general healthy vs non-healthy classification~\cite{szafranowska2022sharing}. It works at patch level, classifying regions-of-interest extracted as grayscale patches with pixel dimension of 224x224 stacked across 3 channels. Pixel values were normalised to fall within the range of 0 to 1. As classification model, a ResNet50~\cite{resnet} was used, which we initialise with weights pretrained on the ImageNet~\cite{imagenet} dataset. To optimise the training process, only the parameters of the last layer were kept trainable. For the multiclass task, there were finally 6147 trainable parameters.

\begin{figure}[h]
\centering
\fbox{\includegraphics[scale=0.37]{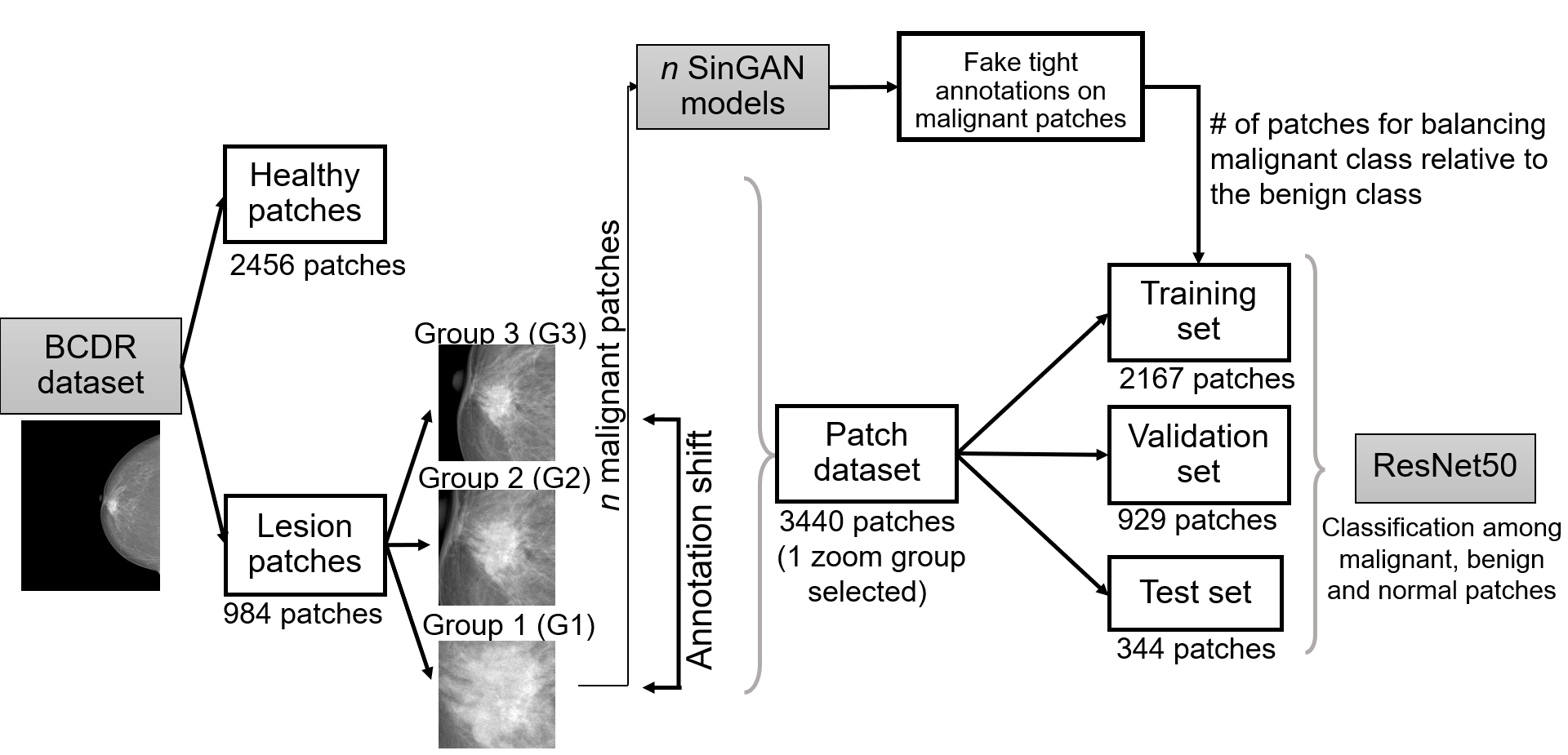}}
\caption{General pipeline of experiments. Patches from healthy and lesion samples are extracted from the BCDR dataset~\cite{bcdr_dataset}, with three patches at distinct zoom levels (G1, G2, G3) for each lesion. The dataset is randomly split into training, validation, and test sets using three folds for all experiments, ensuring that images from the same patient are in a single set. To augment the malignant class, patches from G1 are selected to train different SinGAN models individually, each on a distinct selected patch. Once trained, a synthetic dataset is created assembling generated samples from \textit{n} SinGAN models. This synthetic dataset is incorporated to balance the training dataset.} 
\label{graph}
\end{figure}

\subsection{Simulating annotation shift}

To simulate annotation shift, we extract patches from the lesions from more and less tightly fitting bounding boxes surrounding the lesion, i.e., with different levels of zoom. In practice, an accurate lesion delineation allows to extract a tight lesion bounding box. On the other hand, rectangle lesion annotations (e.g. performed either by human experts or by object detection models~\cite{garrucho2022domain, garrucho2023high, salim2020external}) contain varying amounts of healthy tissue surrounding the lesions. Therefore, our bounding boxes --- extracted based on different zoom levels --- simulate varying annotation protocols (annotation shift) and thus allow to measure their influence on classification performance.

\begin{figure}[h]
\centering
\includegraphics[scale=0.42]{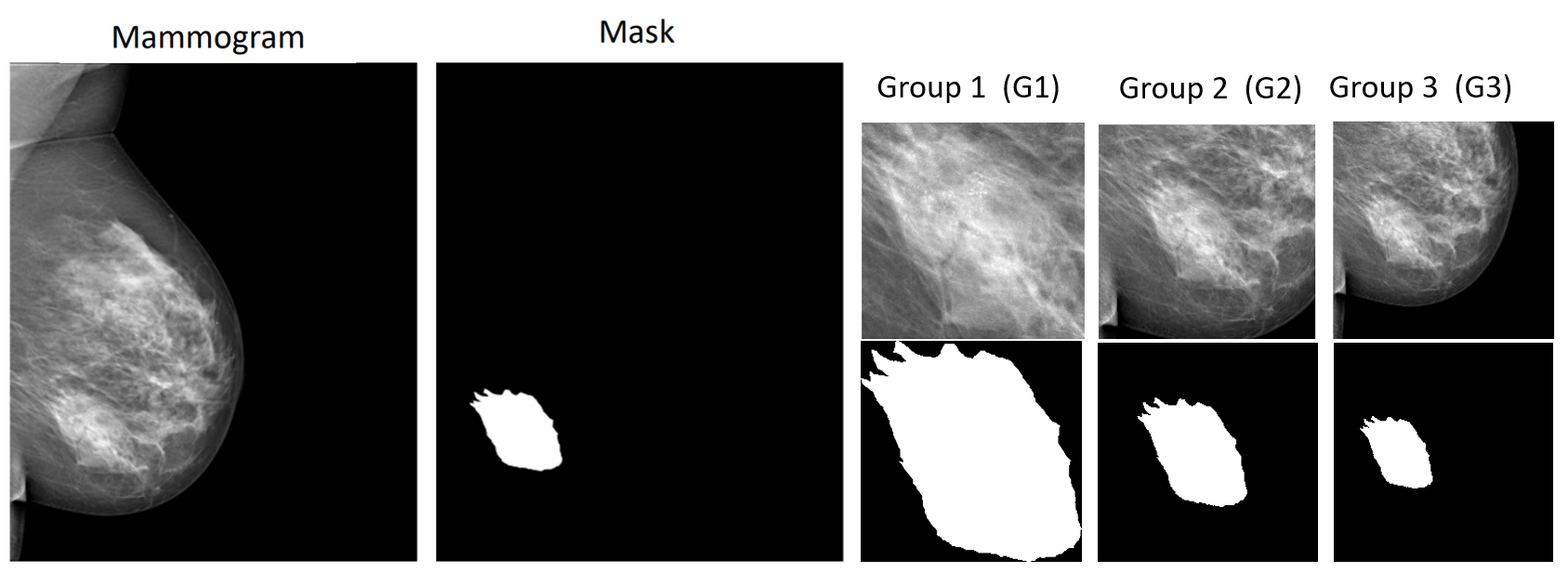}
\caption{Digital mammogram with a biopsy-proven malignant lesion and its corresponding lesion annotation mask. From the third to the fifth column, the extracted patches are depicted ranging from region-of-interest zoom level group 1 (G1) to group 2 (G2) and 3 (G3) with increasing extend of non-lesion tissue visible on the patch.}
\label{fig:healthy_mask}
\end{figure}

Thus, for each lesion, three patches are defined and extracted with different levels of zoom, capturing varying percentages of adjacent healthy tissue. Group 1 (G1) patches correspond to the most accurate bounding box defined around the original annotated lesion delineation mask. Group 2 (G2) and 3 (G3) capture patches with double (200\%) and triple (300\%) the height and width of the original bounding box, respectively. Figure ~\ref{fig:healthy_mask} exemplifies the aforementioned lesion patches extracted from different zoom levels from a digital mammogram. A total of 3440 such patches are extracted, of which 1406 originated from film-scanned and 2034 from digital mammograms. Among these, 2456 patches are from normal mammograms, 699 from mammograms with benign biopsy results, and 285 from mammograms with malignant biopsy results.

\subsection{Single image generative model}

Given its potential and its successes in medical image synthesis \cite{xu2023generative, salle2023cross,thambawita2022singan,medigan,data_synthesis_review} and, particularly, lesion region-of-interest generation~\cite{thambawita2022singan}, we adopt the SinGAN architecture~\cite{SinGAN} to generate multiple synthetic images from a single training image. 
SinGAN is a multi-scale generative adversarial network (GAN)~\cite{goodfellow2014generative} containing a generator-discriminator pair in each scale. The input in each such scale $s_{n}$ consists of a vector $z_{n}$ drawn from a noise distribution $P_{z}$ which, if applicable, is added to the image $x_{n+1}$ generated in the previous scale $s_{n+1}$ to generate image $x_{n}$. Through such iterative GAN-based upsampling, the generator in the last scale $s_{0}$ concludes by outputting a synthetic image $x_{0}$ of the size of the original training image. 

In summary, SinGAN is a stacked ensemble of $\mathcal{N}$ GANs that operate on different scales of the same image, so each scale $n$ has its own generator ($G_n$) and discriminator ($D_n$). Despite the requirement of only a single image as training input, the training process involves sequentially training $\mathcal{N}$ GANs, which is the trade-off for leveraging the capabilities of SinGAN.The training loss for each GAN in each scale incorporates two terms (\ref{eq:loss_singan}): the adversarial loss ($\mathcal{L}_{adv}$) and the reconstruction loss ($\mathcal{L}_{rec}$). 

\begin{equation}
    \min_{G_n}\max_{D_n}\mathcal{L}_{adv}(G_n, D_n) + \alpha\mathcal{L}_{rec}(G_n)
\label{eq:loss_singan}
\end{equation}

The adversarial loss penalises the difference between the distribution of patches in the real image $x_n$ and the distribution of patches in the generated samples $\tilde{x}_n$. This loss term helps the generator to produce samples that resemble the real data distribution. The Wasserstein GAN with Gradient Penalty (WGAN-GP) loss function~\cite{wGAN}, is used for training the generator and discriminator. It was observed that this loss function, compared to other GAN variants, enhances training stability~\cite{SinGAN}. The expression of the WGAN-GP loss function is:
\begin{equation}
    \mathcal{L}=\mathbb{E}_{\tilde{\textbf{x}}\sim \mathbb{P}_g}[f(\tilde{\textbf{x}})] - \mathbb{E}_{{\textbf{x}\sim \mathbb{P}_r}}[f(\textbf{x})] + \lambda\mathbb{E}_{\hat{\textbf{x}}\sim \mathbb{P}_{\hat{x}}}[(||\Delta_{\hat{x}}f(\hat{\textbf{x}})||_2-1)^2] 
\end{equation}
\label{eq:WGAN-GP-eq}

In Equation \ref{eq:WGAN-GP-eq}, the first two terms are the original WGAN loss and the right term is the gradient penalty with $\lambda$ the penalty coefficient.

\begin{figure}[h]
    \centering
    \fbox{\includegraphics[width=0.9\linewidth]{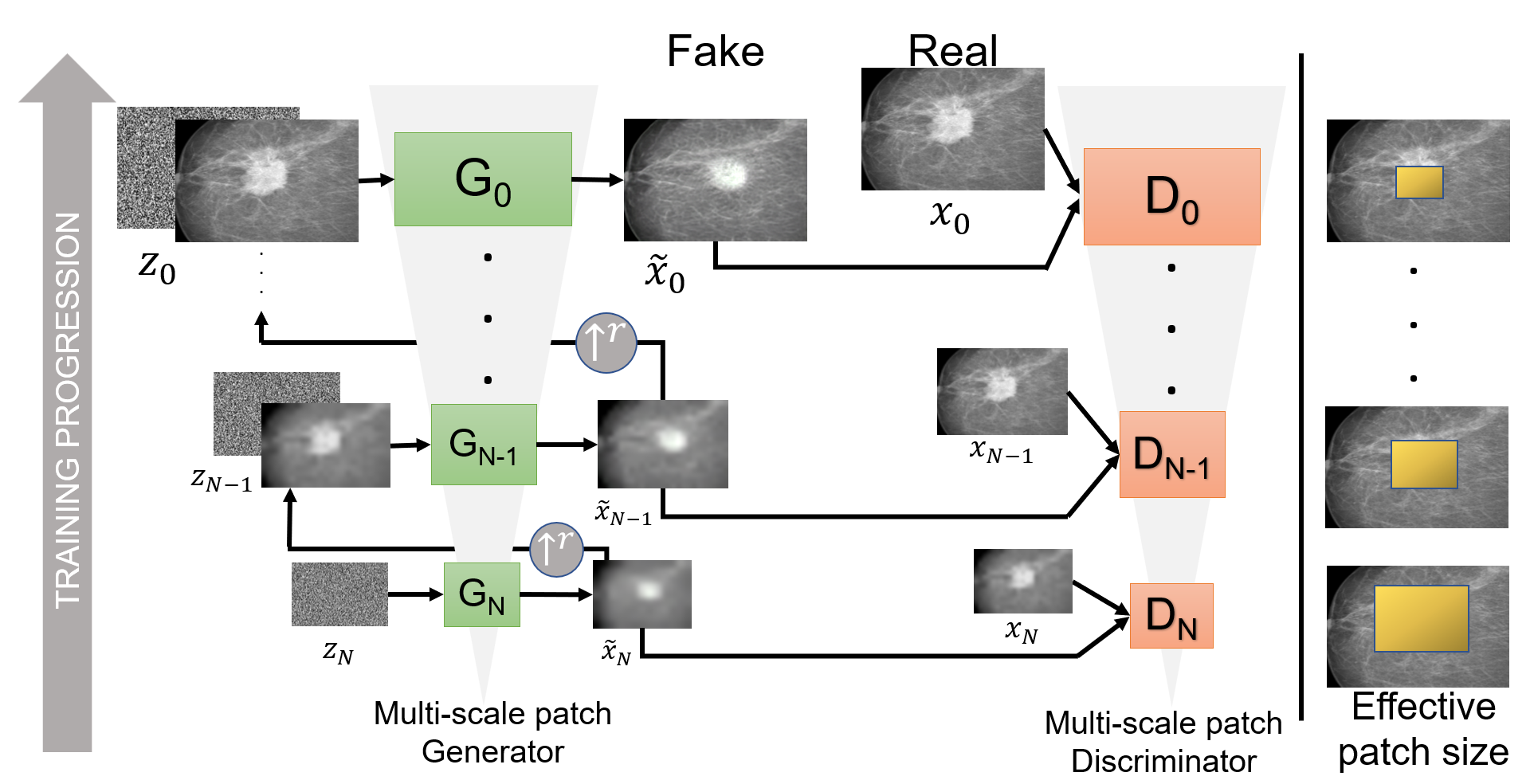}}
  \caption{Implemented pipeline of the SinGAN framework adopted from 
  Shaham et al\cite{SinGAN}. One GAN operates at each of the $n$ different SinGAN image scales. The training process starts with the coarsest scale and progresses to the finest scale. Each GAN in the hierarchy learns to generate realistic images at its respective scale, capturing both global and local details. At each scale $s_{n}$, the image from the previous scale, $\tilde{x}_{n+1}$, is upsampled and added to the input noise map, $z_n$. The result is fed into the generator ($G_n$), whose output is the residual image $\tilde{x}_{n}$ of scale $s_{n}$ fed into discriminator $D_n$ during training and passed to the next scale during inference. 
  }
  \label{fig:graph_singan}
\end{figure}

We implement SinGAN based on the official code repository~\cite{SinGAN} with adaptations for hyperparameter setup. A scale factor of 0.8 was chosen, which means that the resolution of the image when passing it to the next scale is reduced by 20\%. For each scale, the final model after 30 training epochs was selected. The default values were used for the rest of the hyperparameters. The framework of the SinGAN architecture is depicted in Figure~\ref{fig:graph_singan}, illustrating this hierarchical structure of the model. 

\subsection{Addressing annotation shift with synthetic data}

Apart from annotation shift, our study addresses the challenge of an imbalanced dataset with biopsy-proven malignant breast lesions as minority class impacting classification performance as further elaborated in Section~\ref{sec:results}. 
To address this issue, we propose and empirically evaluate the usage of synthetic data generated by single-image generative models to augment the malignant samples. Therefore, our approach is two-fold. Firstly (i), we mitigate annotation shift by generating single image variations of the target annotation distribution controlled via bounding box zoom level. Secondly (ii), we further reduce the effect of dataset imbalance by balancing the dataset with synthetic images corresponding to the minority class. 

We train our SinGAN models exclusively on malignant lesion patches from a specific level of zoom (e.g. G1). We note that G1 patches represent the most accurate lesion annotations and, thus, are the most challenging to obtain. For instance, G2 or G3 lesion patches can be retrieved from a G1 annotation, but not vice versa, as the lesion of a G2 or G3 annotation can be (partly) outside the boundaries of a cropped (zoomed-in) G1 patch.

To study the impact of using multiple SinGANs to generate a more diverse synthetic dataset, we individually train multiple SinGAN models, each on a distinct single malignant lesion image. Next, synthetic datasets are assembled, where each SinGAN contributes the same number of samples, which, put together, balance the training dataset of the classification model (i.e. same number of benign and malignant lesions). The synthetic datasets (based on a varying number of SinGANs) is used during classifier training for data augmentation. This approach allows us to systematically study the impact of using different numbers of SinGAN models in generating the synthetic dataset. 

Alternatively, as baseline for experiments without synthetic samples, a weighted random sampling technique is applied to avoid class imbalance to bias the classification model. This technique assigns a sampling probability based on class frequency to each sample in the classifier's training and validation set. For experiments with synthetic data, weighted random sampling was only applied to the validation set. 

\section{EXPERIMENTS AND RESULTS}\label{sec:results}
In this study, we initially assessed a binary classification task distinguishing between healthy and lesion-containing patches, yielding a test accuracy of 0.924 ± 0.009 and a test ROC-AUC (Area under the Receiver Operating Characteristic Curve) of 0.971 ± 0.009, indicating how effectively the model classifies this binary class. Subsequent experiments described in this section extended this setup to multiclass classification, classifying patches either as healthy, benign, malignant. This extended approach enables a comprehensive analysis of detected abnormalities, aiming to predict biopsy outcomes. 

Each experiment ran for 100 epochs and the model from the epoch with the best validation loss was selected. Models were evaluated based on a train-validation-test split across three folds, ensuring that each patient was present in only one of the sets. The data was partitioned into 10\% for testing (344 samples), 63\% for training (2167 samples), and 27\% for validation (929 samples). The overall experimental process is represented in Figure~\ref{graph}. The experiments were conducted with consistent hyperparameters to ensure fair comparisons between methods. These included a fixed batch size of 128, utilizing the adaptive moment optimiser (Adam~\cite{kingma2014adam}) with default beta parameters ($\beta_1$=0.9 and $\beta_2$=0.999), and employing a learning rate scheduler that gradually decayed the learning rate which started at $10^{-2}$. The scheduler had a step of 5 epochs and a gamma value of 0.1. For both the binary classification problem and the multiclass task, a binary cross-entropy loss function was employed. All experiments were run on a NVIDIA RTX 2080 Super 8GB GPU using the PyTorch library~\cite{pytorchPaszke}. Training the classifier for 100 epochs took approximately 3 hours in this setup.

\subsection{Analysing model generalisation across all annotation variations}
\label{sec:exp1}
Simulating annotation shift, each experiment involved training the classifier on images from one annotation protocol (represented by a zoom group) and testing on samples from all zoom groups. As upper bound comparison, an additional model was trained using lesion patches from all three zoom groups. Table~\ref{fig:experiment-accuracies-tab1} summarises the empirical results measured using test set classification accuracy and ROC-AUC reported alongside their standard deviation across 3 folds. The ROC-AUC metric facilitates a thorough examination of the separability of the classes across all possible thresholds. For the three-class task, an One-vs-Rest (OvR) strategy for reporting this metric was used.

It is demonstrated that training on lesions from all three groups yields the highest accuracy and ROC-AUC for all classes. This indicates that if there are several annotations available for one sample, it is beneficial for model robustness to use all of these annotations during training. 

Observing the ROC-AUC across classes, training on only G2 outperformed exclusive G1 or G3 training, indicating that G2 captures essential features for testing on both extremes. If a range of annotation options exist it can, hence, be optimal to choose the one closest to the average across annotations to generalise best across all annotation options. Conversely, exclusive G1 or G3 training results in poorer performance, emphasising the impact of annotation shift on classifier performance. Overall, the zoom level substantially influenced classifier performance across zoom levels. The model's dependence on precise annotations is highlighted, which is prone to affect generalisation in real-world computer-aided diagnosis applications, especially in cases where annotation protocols vary.

\begin{table}[ht]
  \centering
  \begin{tabular}{|c|c|c|c|c|c|}
  \hline
    \multicolumn{2}{|c|}{\textbf{Experiment}} & \multirow{2}{*}{\textbf{Overall accuracy} (Mean$\pm$SD)} & \multicolumn{3}{|c|}{\textbf{ROC-AUC per class} (Mean$\pm$SD)} \\
    \cline{1-2}\cline{4-6}
    Train-Val & Test & & Healthy & Malignant & Benign\\
    \hline
    G1:G2:G3 & G1:G2:G3 & $\mathbf{0.837\pm0.045}$ & $\mathbf{0.977\pm0.009}$ & $\mathbf{0.901\pm0.037}$& $\mathbf{0.942\pm0.021}$\\
    G1 & G1:G2:G3 & $0.748\pm0.044$ & $0.938\pm0.022$ & $0.787\pm0.046$& $0.931\pm0.027$\\
    G2 & G1:G2:G3 & $0.804\pm0.035$ & $0.969\pm0.008$ & $0.875\pm0.046$& $0.938\pm0.017$\\
    G3 & G1:G2:G3 & $0.807\pm0.035$ & $0.968\pm0.008$ & $0.804\pm0.055$& $0.927\pm0.021$\\ 
    \hline
  \end{tabular}
   \caption{Mean classification accuracy and ROC-AUC for the three classes alongside their standard deviations (SD) across 3 runs. While trained on only on one group of zoom level (G1, G2, G3 in row 2, 3, 4, respectively), the models are tested on samples from all the groups (G1, G1, G3) The upper-bound baseline model in row 1 is trained and tested on all zoom level groups.}
   \label{fig:experiment-accuracies-tab1}
\end{table}

\subsection{Assessing cancer classification robustness under specific annotation shifts}
\label{sec:exp2}

\begin{figure}[h]
    \centering
    \includegraphics[scale=0.55]{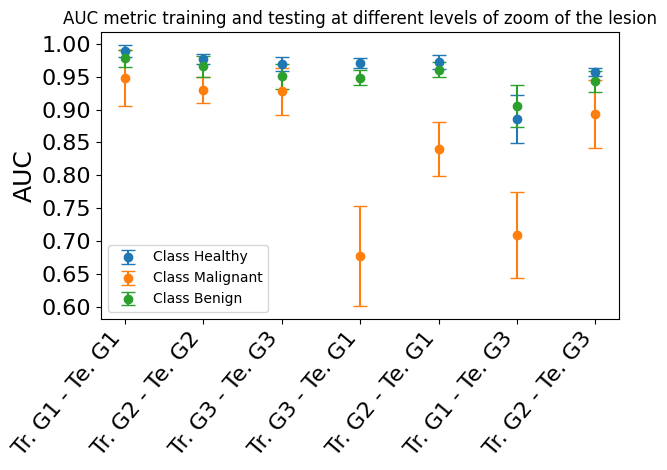}
    \caption{Area under the Receiver Operating Characteristic (ROC) Curve  for each class computed using the One-vs-Rest (OvR) strategy for training and testing on specific varying zoom levels. The vertical bars in the plot representing the standard deviation.  The results reveal varying model robustness under annotation shift.}
    \label{fig:AUC_plot_different_zoom}
\end{figure}

Focusing on the classification class of interest (i.e., 'Malignant') to indicate the presence of a cancerous lesion, we assess classifier performance for training and testing on specific varying zoom levels.
As expected, the highest accuracy and ROC-AUC was achieved when testing on lesion patches from the same zoom level group as the training data. On the contrary, the lowest accuracy occurred when the classifier trained on group 1 (G1) was tested on group 3 (G3), which also corresponded to the second lowest ROC-AUC result. The lowest ROC-AUC was observed when the classifier trained on group 3 was tested on group 1. With results summarised in Table ~\ref{fig:experiment-accuracies-tab2} and in Figure~\ref{fig:AUC_plot_different_zoom}, 
we note that training on the intermediate zoom level of group 2 yields better performance for the malignant class when testing on groups 1 or 3 compared to training on those groups. Overall, it is demonstrated that annotation shift considerably impacts classification performance and further measures to ensure model robustness in such scenarios are needed.

\begin{table}[ht]
  \centering
  \begin{tabular}{|c|c|c|c|}
  \hline
    \multicolumn{2}{|c|}{\textbf{Experiment}} & \textbf{Overall accuracy} & \textbf{ROC-AUC: Malignant class} \\
    \cline{1-2}
    Train-Val & Test & (Mean $\pm$ SD) & (Mean $\pm$ SD) \\
    \hline
    G1 & G1 & $\mathbf{0.929\pm0.034}$ & $\mathbf{0.948\pm0.043}$\\
    G2 & G2 & $0.899\pm0.024$ & $0.930\pm0.020$\\
    G3 & G3 & $0.887\pm0.031$ & $0.928\pm0.036$\\ 
    G3 & G1 & $0.865\pm0.088$ & $0.677\pm0.076$\\
    G2 & G1 & $0.877\pm0.009$ & $0.840\pm0.041$\\ 
    G1 & G3 & $0.780\pm0.021$ & $0.709\pm0.066$\\
    G2 & G3 & $0.835\pm0.025$ & $0.893\pm0.052$\\
    \hline
  \end{tabular}
   \caption{Results indicating model robustness under annotation shift. Accuracy accross classes and ROC-AUC metrics for the class of interest ('Malignant') at testing time are reported alongside their standard deviation (SD) across three folds. The ROC-AUC metrics for the three classes in these experiments are plotted in Figure \ref{fig:AUC_plot_different_zoom}.
   }
   \label{fig:experiment-accuracies-tab2}
\end{table}

\subsection{Enhancing model robustness with SinGAN-based data augmentation}
\label{sec:singan-augmentation}
\begin{figure}[h]
    \centering
    \includegraphics[scale=0.6]{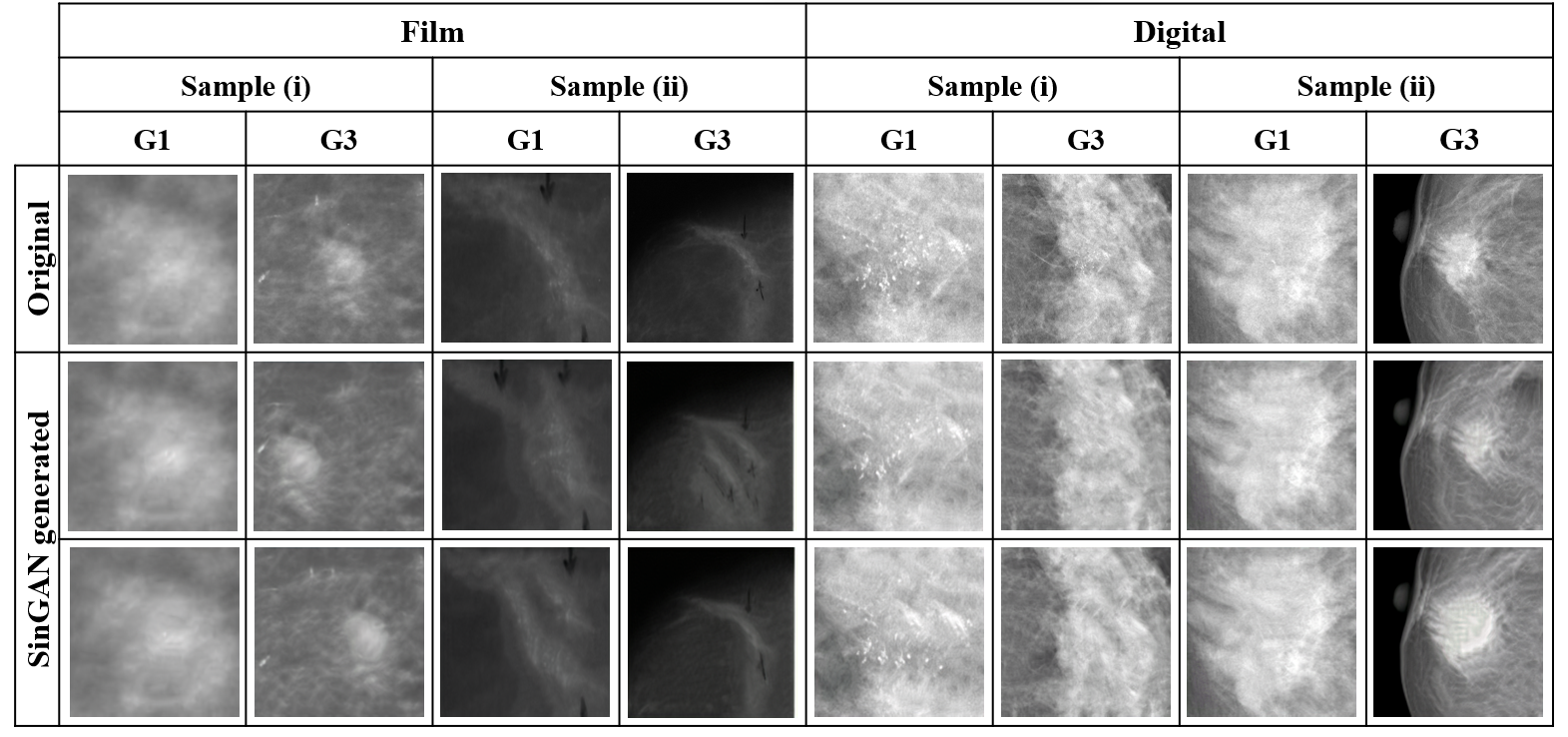}
    \caption{The figure presents the original samples utilized for training each SinGAN model, alongside two synthetic samples generated by the respective trained model. There are two samples for each format (film and digital), and for each sample, its corresponding patch from zoom groups G1 and G3. For augmenting the dataset only samples generated from models trained on a sample from group 1 (G1) were used. This presentation aims to demonstrate the realism achieved by the SinGAN models in generating synthetic content.}
    \label{fig:singan_samples}
\end{figure}

In this set of experiments, we investigate data augmentation using SinGAN models to enhance classifier performance under annotation shift for the 'malignant' class. Not only showed this class consistently the lowest classification performance in previous experiments, but it was also heavily impacted by annotation shift, as shown in Table~\ref{fig:experiment-accuracies}. 
For each experiment, the classifier was trained on zoom group 3 (G3). Each of the SinGAN models used for generating new data was trained on a different malignant (G1) lesion image. The malignant lesion images used to train each of the SinGAN models, along with two examples generated with each of them, are shown in Table \ref{fig:singan_samples}.

\begin{figure}[h]
    \centering
    \includegraphics[scale=0.5]{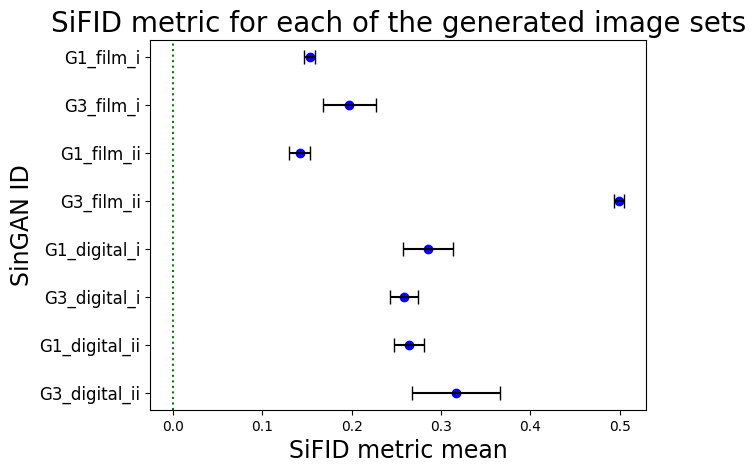}
    \caption{SiFID metric values distribution is presented for datasets generated by SinGAN models trained on different single images. The y-axis represents the ID of the training image used to generate each dataset, indicating the group of zoom level (G1 or G3) and the format (digital or film). The dots represent the mean value, and the horizontal black line represents the standard deviation. For reference, the metric was computed with respect to a dataset of random noise images as an upperbound, which had a SiFID of $39.282\pm 0.143$, so the metrics obtained demonstrate the excellent quality of the generated images. The corresponding original images and the generated ones can be found in Figure \ref{fig:singan_samples}.}
    \label{fig:sifid_plot}
\end{figure}

To quantitatively asses synthetic data quality, we used the Single Image FID (SiFID) metric~\cite{SinGAN}  measuring the similarity between the real and synthetic images. The SiFID values obtained are presented in Figure \ref{fig:sifid_plot}. To establish a benchmark, these metrics were compared against an upper bound calculated from a dataset of random noise images, resulting in a SiFID of $39.282\pm 0.143$. The SiFID metrics, ranging from 0.14 to 0.49, illustrate that the synthetic images closely match the feature distributions of real images. This alignment indicates a high level of quality and fidelity in the process of image generation.

\begin{figure}[h]
    \centering
    \includegraphics[scale=0.54]{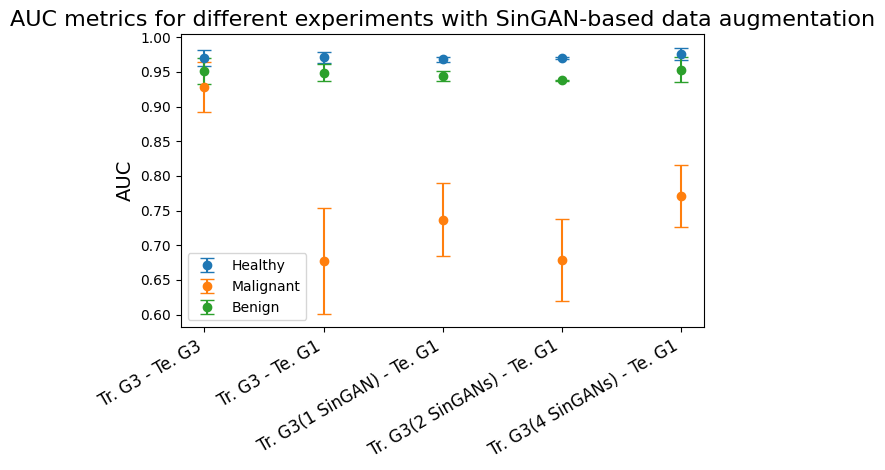}
    \caption{ROC-AUC 
    for each class across experiments where the classifier 
    was trained on zoom group 3 (G3), optionally augmented with synthetic SinGAN generated training images balancing the training dataset.     
    Synthetic images are generated by 1, 2, or 4 SinGAN models trained on single images that had been randomly sampled from the G1 zoom group and result in a performance increase for the malignant class.}
    \label{fig:AUC_plot_different_numberSinGAN}
\end{figure}

In each experiment, synthetic data were generated from 1, 2, and 4 different SinGAN models, each trained on distinct single malignant lesion images. The number of total generated samples is the same for all the experiments. This approach allowed for the assessment of the impact of augmented data diversity. The OvR ROC-AUC scores per class for each experiment are presented in Table \ref{fig:AUC_table_different_numberSinGAN} and Figure \ref{fig:AUC_plot_different_numberSinGAN}. The most favorable outcome for the malignant class in terms of ROC-AUC was achieved when generating new data from 4 different SinGAN models. Despite using as few as 1 to 4 lesion images with in-domain annotations, SinGAN data augmentation improves upon the baseline for the underrepresented malignant class.

Additionally, an interesting observation was made: In the experiment utilizing 2 SinGAN models, a performance degradation was observed compared to using either 1 or 4 models. This suggests that the selection of these few lesion images accurately annotated for training SinGAN has a direct impact on the performance enhancement of the classifier. The choice of images used to train the SinGAN model and, therefore, the data employed for data augmentation, likely substantially influences the classifier's performance. We are currently developing a method to carefully select the most suitable images for data augmentation, aiming to refine the process and enhance classification outcomes.

\begin{table}[ht]
\centering
\begin{tabular}{ll|llll|}
\cline{3-6}
 & & \multicolumn{4}{c|}{\textbf{Number of SinGAN models used for data augmentation}}\\ \cline{3-6} 
 & \multicolumn{1}{c|}{\textbf{}} & \multicolumn{1}{c|}{\begin{tabular}[c]{@{}c@{}}No synthetic\\data generated\end{tabular}} & \multicolumn{1}{c|}{1 SinGAN model} & \multicolumn{1}{c|}{2 SinGANs models} & \multicolumn{1}{c|}{4 SinGAN models} \\ \hline
\multicolumn{1}{|l|}{\multirow{3}{*}{\rotatebox{90}{\textbf{Class}}}} & Healthy & \multicolumn{1}{c|}{$0.971 \pm 0.008$}    & \multicolumn{1}{c|}{$0.968 \pm 0.004$}         & \multicolumn{1}{c|}{$0.970 \pm 0.001$}           & $\mathbf{0.976 \pm 0.009}$                      \\ \cline{2-2} 
\multicolumn{1}{|c|}{} & Malign & \multicolumn{1}{l|}{$0.677 \pm 0.076$}    & \multicolumn{1}{c|}{$0.737 \pm 0.053$}         & \multicolumn{1}{c|}{$0.679 \pm 0.059$}           & $\mathbf{0.771 \pm 0.045}$ \\ \cline{2-2} 
\multicolumn{1}{|c|}{} & Benign & \multicolumn{1}{c|}{$0.949 \pm 0.012$}    & \multicolumn{1}{c|}{$0.944 \pm 0.007$}         & \multicolumn{1}{c|}{$0.938 \pm 0.001$}           & $\mathbf{0.953 \pm 0.018}$ \\ \hline
\end{tabular}
\caption{Table displays experiment results using synthetic data generated from 1, 2, and 4 SinGAN models trained on distinct malignant lesion images from zoom level G1, known for their accurate annotations. Total generated samples remain consistent across experiments. Original classifier training samples are from zoom level G3, augmented with SinGAN-generated data. OvR ROC-AUC scores for classifier testing on G1 samples per class presented, evaluating data diversity impact. SinGAN augmentation enhances underrepresented malignant class; however, using only 2 SinGAN models leads to performance degradation, highlighting the importance of accurately annotated lesion images for SinGAN training. Original malignant lesion images used for SinGAN training are listed in Table ~\ref{fig:singan_samples}.}
\label{fig:AUC_table_different_numberSinGAN}
\end{table}

\pagebreak

\subsection{Combination with traditional data augmentation techniques}
Experiments were conducted to compare data augmentation using SinGAN-generated samples with a traditional data augmentation method, specifically single-image (G1) oversampling. The experiments with SinGAN-augmented data involved generating a synthetic dataset from 4 SinGAN models trained on different lesion images from group G1, as it yielded the best results in section ~\ref{sec:singan-augmentation}. For consistency, the same four training images utilized for training each SinGAN model were also employed for the oversampling method. Additionally, experiments were repeated with a random resized cropping applied to the test samples. This random resized cropping allows us to additionally measure a classification robustness in the case where the lesion is not at the center of the region-of-interest image. As the assumption of a centered lesion likely does not hold in clinical settings, we analyse whether SinGAN's capability of varying lesion position observable in Figure \ref{fig:singan_samples} is beneficial for robustness in our downstream lesion classification task.

The classifier was trained, as in the experiments of section~\ref{sec:singan-augmentation}, on samples from group zoom G3, while the data augmentation was performed with samples from group zoom G1. Furthermore, the classifier was evaluated on samples from both zoom groups G1, characterized by accurate lesion annotation with minimal healthy tissue inclusion in the patch, and G3, featuring less zoom to the lesion.

In all experiments, an ensemble architecture was utilized. For both augmentation techniques, SinGAN-based and with oversampling, and for the baseline (with no data augmentation), three models were trained with a train-validation split across three folds, ensuring that each patient was present in only one of the sets. For evaluation, a fixed test set comprising images from patients exclusively allocated to this set was utilized. The ensemble prediction was determined as the mean of the predictions generated by each of the three models within the ensemble. In the case of the ensemble combining SinGAN augmentation and oversampling, a total of six models were configured for the ensemble. It is important to note that the test set was not altered across any of the described experiments. The test samples are sourced from patients who are not represented in the validation or training sets.

While SinGAN-based results show improvement compared to the baseline without data augmentation, they are not substantially superior to our implemented traditional oversampling method indicating further potential of the latter. To this end, we further extend our method to an ensemble architecture combining both SinGAN and oversampling methods. Respective experiments demonstrated that our ensemble further increased classification performance, e.g., compared to applying each method independently. This suggests that SinGAN-based annotation shift mitigation can be further enhanced when used in combination with other methods such as traditional data augmentation techniques. 

\begin{table}[h]
\centering
\begin{tabular}{|c|c|c|c|c|c|}
\hline
\multicolumn{6}{|c|}{\textbf{ROC-AUC (Mean $\pm$ SD) of the Malignant class for data augmentation techniques comparison}}  \\
\hline
\multicolumn{2}{|c|}{Experiment configuration} & \multicolumn{4}{|c|}{
Data augmentation method}\\
\hline
Test group & Random resized crop & Baseline & SinGAN & Oversampling & SinGAN and oversampling \\
\hline
G1 & No & 0.636 &  0.667 &  \textbf{0.708} & 0.692 \\
G1 & Yes & 0.404 & 0.623 &  0.637 &  \textbf{0.666}\\
G3 & No &  0.917 &  \textbf{0.942} & 0.915 &  \textbf{0.942} \\
G3 & Yes &  0.777 & 0.816 &  0.831 & \textbf{0.841} \\
\hline
\end{tabular}
\caption{ROC-AUC of malignant class comparing SinGAN-generated sample augmentation with traditional single-image (G1) oversampling. Classifier trained on G3 samples, augmented with G1 samples, and evaluated on both G1 and G3 samples. Experiments were also done including random resized crop on test samples to simulate annotation shift impact. An ensemble architecture was employed for all experiments, comprising three models for each augmentation technique and six models for the combined ensemble. Evaluation was conducted on a fixed test set with patients exclusively allocated to this set, ensuring consistent testing conditions across all experiments.}
\end{table}

\section{DISCUSSION and CONCLUSION}

In this study, we focused on the development of a high-accuracy malignancy classification model trained to distinguish cancerous from benign breast lesions. Our investigation uncovered several critical insights into the challenges and potential solutions in this domain. Firstly, we quantified the impact of annotation shift on classification performance through simulations with varying bounding box tightness representing different annotation protocols. To simulate annotation shift, patches from the lesions were extracted from more and less tightly fitting bounding boxes surrounding the lesion, i.e., with different levels of zoom. Our experiments on sections ~\ref{sec:exp1} and ~\ref{sec:exp2} revealed that the zoom level or annotation accuracy substantially influenced classifier performance, emphasizing the model's dependence on precise annotations, which could affect generalization in real-world computer-aided diagnosis applications where annotation protocols vary.

Moreover, we observed a performance disparity in the classifier's prediction of malignant cases compared to benign and healthy images. This class consistently exhibited the lowest classification performance and was heavily impacted by annotation shift. To address this, we explored data augmentation techniques, particularly focusing on enhancing performance for the 'malignant' class. Our explorations with synthetic data from single-image generative AI models demonstrated the potential to enhance performance with as few as 1 to 4 in-domain annotations. However, we also found that the choice of images used to train the SinGAN model and, therefore, the data employed for data augmentation, significantly influences the classifier’s performance. We are currently working on refining a method for selecting the optimal images for the task of data augmentation, aiming to further improve classification outcomes.

Additionally, we conducted a comparative analysis between SinGAN models and a traditional augmentation technique. Specifically, we employed a traditional oversampling method by augmenting the training dataset with four accurately annotated malignant images. In parallel, four SinGAN models were individually trained on each of these images to generate and augment the data. Our observations revealed that while SinGAN-based augmentation showed improvement in the classifier performance compared to the baseline without augmentation, it did not exhibit substantial superiority over our traditional oversampling approach, indicating the continued potential of the latter. To further explore, we conducted an extended experiment where we combined both methods in an ensemble architecture. This experiment demonstrated that SinGAN models could enhance traditional data augmentation techniques when integrated into an ensemble architecture.

In conclusion, our study reveals multiple insights into breast cancer risk assessment using deep-learning classification models. These findings shed light into the potential of one-shot generative AI models in mitigating annotation shift challenges and collectively contribute to advancing computer-assisted breast cancer diagnosis and detection.

\acknowledgments 
 

This project has received funding from the European Union’s Horizon Europe and Horizon 2020 research and innovation programme under grant agreement No 101057699 (RadioVal) and No 952103 (EuCanImage), respectively. Also, this work was partially supported by the project FUTURE-ES (PID2021-126724OB-I00) from the Ministry of Science and Innovation of Spain.

\pagebreak
\bibliography{report} 
\bibliographystyle{spiebib} 

\end{document}